\documentclass[conference, a4paper]{IEEEtran}
\usepackage{booktabs}
\usepackage{multirow}
\usepackage[vlined,linesnumbered,titlenotnumbered,ruled]{algorithm2e}

\usepackage[fleqn]{amsmath}
\usepackage{amsfonts}
\usepackage{amssymb}
\usepackage{graphicx}
\usepackage{gensymb}

\usepackage[flushleft]{threeparttable}

\usepackage{xcolor}

\ifCLASSOPTIONcompsoc
\usepackage[nocompress]{cite}
\else
\usepackage{cite}
\fi
\begin{document}
	
\title{{An Efficient Software-Hardware Design Framework for Spiking Neural Network Systems}}
\IEEEpubidadjcol
\author{\IEEEauthorblockN{Khanh N. Dang\IEEEauthorrefmark{1}, and Abderazek Ben Abdallah\IEEEauthorrefmark{2}}
	\IEEEauthorblockA{\IEEEauthorrefmark{1}SISLAB, VNU University of Engineering and Technology, Vietnam National University, Hanoi, Hanoi, 123106, Vietnam \\}
	
	\IEEEauthorblockA{\IEEEauthorrefmark{1}\IEEEauthorrefmark{2}Adaptive Systems Laboratory, The University of Aizu, 
		Aizu-Wakamatsu, Fukushima 965-8580, Japan.}
	
	Email: \IEEEauthorrefmark{1}khanh.n.dang@vnu.edu.vn; \IEEEauthorrefmark{2}benab@u-aizu.ac.jp}


%
%

\IEEEoverridecommandlockouts\IEEEpubid{\makebox[\columnwidth]{ 978-1-7281-5184-7/19/\$31.00~ \copyright~2019~IEEE} \hspace{\columnsep}\makebox[\columnwidth]{ }}
\maketitle

\IEEEpeerreviewmaketitle
\begin{abstract}
	
Spiking Neural Network (SNN) is the third generation of Neural Network (NN) mimicking the natural behavior of the brain. By processing based on binary input/output, SNNs offer lower complexity, higher density and lower power consumption. 
This work presents an efficient software-hardware design framework for developing SNN systems in hardware. In addition, a design of low-cost neurosynaptic core is presented based on packet-switching communication approach.
The evaluation results show that the ANN to SNN conversion method with the size 784:1200:1200:10 performs 99\% accuracy for MNIST while the unsupervised STDP archives 89\% with the size 784:400 with recurrent connections. The design of 256-neurons and 65k synapses is also implemented in ASIC 45nm technology with an area cost of 0.205 $m m^2$. 

 
\end{abstract}

\IEEEpeerreviewmaketitle
\section{Introduction}
Brain-inspired computing or neuromorphic computing is the next generation of artificial intelligence that extend to areas of human cognition. As first introduced by Carver Mead in 1990~\cite{Mead1990Neuromorphic}, Very Large Scale Integration (VLSI) with the help analog components could mimic the behavior of brains with low costs. The computational building blocks are the replicated version of neuron which receives, process and sends possible output spike. Recently, several researchers and companies have been investigating to integrate large number of neurons on a single chip while providing efficient and accurate learning~\cite{Davies2018Loihi,Akopyan2015TrueNorth,Furber2014SpiNNaker,Benjamin2014Neurogrid,Schemmel2010wafer}.

Spiking neural network (SNN)~\cite{Vu2019Comprehensive,Vu2019Fault}  is a novel model for arranging the replicated neurons to emulate natural neural networks that exist in biological brains.
Each neuron in the SNN can fire independently of the others, and doing so, it sends pulsed signals to other neurons in the network that directly change the electrical states of those neurons. By encoding information within the signals themselves and their timing, SNNs simulate natural learning processes by dynamically remapping the synapses between artificial neurons in response to stimuli. 
Incoming spikes are integrated in soma to its membrane potential. If the membrane potential crosses the threshold, the neuron sends an outgoing spikes to an axon. The axons send the spike to the downstream neurons.

To provide functional systems for researchers to implement SNNs, several works~\cite{Davies2018Loihi,Akopyan2015TrueNorth,Furber2014SpiNNaker,Benjamin2014Neurogrid,Schemmel2010wafer} have been proposed to provide SNN platform. Different from ANNs, SNNs take consideration the time in their computation. One or several neurons might send out spikes, which are represented by single-bit impulses, to neighbors through connections (synapses). Each neuron has it own state values that decide the internal change and spike time. The network is composed of individual neurons interacting through spikes. 
Incoming spikes go through a synaptic weight storage and is converted to weighted inputs. The membrane potential of the neuron integrates the weighted inputs and causes an outgoing spike (or firing) if it is higher than the threshold. The membrane potential is reset to resting voltage after firing and the neuron falls into refractory mode for several time steps.

Currently, there are two major benefits of using SNNs instead of ANNs: (1) lower complexity and power consumption and (2) early-peek result. Since SNNs is mainly based on multiplications of binary input, there is no actual multiplication module needed. Unlike MAC (multiply-accumulate) unit in ANN, the computation unit of SNN mainly requires adders which significantly reduce the area cost and computation time. 
Also, by representing in binary input, several low-power methods could be used (i.e. clock/power gating, asynchronous communication, and so forth) and thank to their lower complexity, the power consumption is also smaller. 
Second, SNNs could provide an early peek result that provide fast response time. As shown in~\cite{Diehl2016Conversion}, the SNN can be nearly as accurate as after 350 time steps at just after 100 time steps. 
This could help reduces the power consumption by cutting the operation of the unneeded module and obtain fast response time.

On the other hand, \textit{Kim et al.}~\cite{Kim2016Neurocube} have introduced  a TSV-based 3D-IC neuromorphic design that allows less bandwidth utilization by stacking multiple high density memory layers. 
This brings up an opportunity to integrating 3D-ICs for SNNs which requires high density and distributed memory access. While \cite{Kim2016Neurocube} stucks 2D-mesh network, using 3D mesh network is more efficient for computing and communication. A study proposed in \cite{Vu2019Fault, Vu2019Comprehensive} shows that 3D NoCs outperforms the 2D NoCs in bandwidth efficiency and spiking frequency.

Although several SNN architectures have been proposed, there are still several challenges as follows: 
\begin{enumerate}
	\item \textit{Memory technology and organization:} since a large-scale SNN architecture requires an enormous amount of neurons and weights, using an efficient memory technology, organization, and access need to be carefully investigated. While most existing works~\cite{Davies2018Loihi,Akopyan2015TrueNorth,Furber2014SpiNNaker,Benjamin2014Neurogrid,Schemmel2010wafer} rely on serializing using central SRAM to perform computation, we observe distributed memory could significantly improve the performance.
	
\item \textit{On-chip inter-neuron communication} is another issue that needs careful investigation. Since SNNs are communication-intensive where each neuron sends its output spikes to several neurons, the congestion in on-chip communication may occurs. As being implemented in VLSI, large fan-out is not desired due to the lack of signal strength which leads to a large buffer requirement.
	
\item \textit{On-line learning} is also another issue which needs to be properly addressed. Despite of having several benefits, the current learning methodology is not efficient in terms of accuracy and training time. Multiple layer networks seem hard to be trained.		
\end{enumerate}
This work presents a comprehensive software-hardware platform for designing SNNs based on on-line learning algorithm. In addition, this work proposes an architecture, design and evaluation of a low-cost spiking neural network system.
This paper is organized as follows: Section~\ref{sec:arch} shows the proposed SNN architecture. To provide an overview of the proposed platform, Section~\ref{sec:inte} describes how to integrate each architectural module. Section~\ref{sec:eva} evaluates the design and finally Section~\ref{sec:concl} concludes the study.

\section{System Architecture}  \label{sec:arch}
\begin{figure}
	\centering
	\includegraphics[width=1\linewidth]{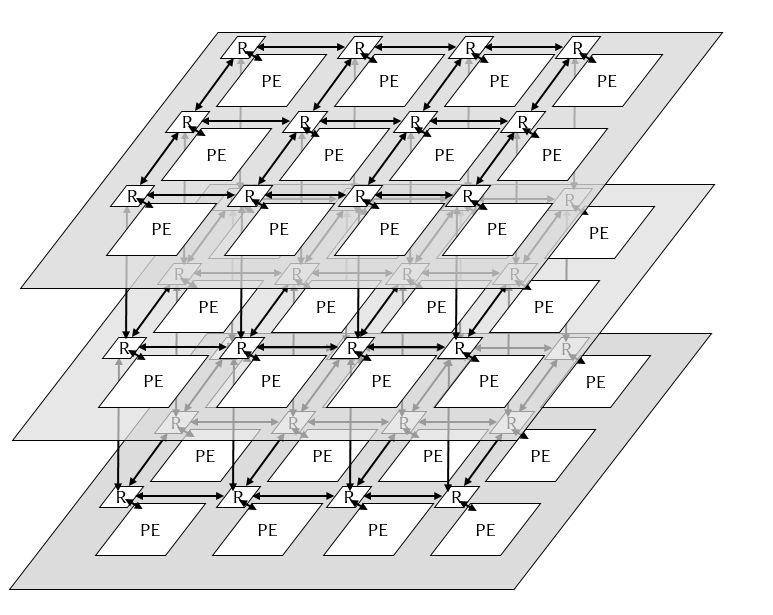}
	\caption{System block diagram.}
	\label{fig:systtop}
\end{figure}

\begin{figure}
	\centering
	\includegraphics[width=1\linewidth]{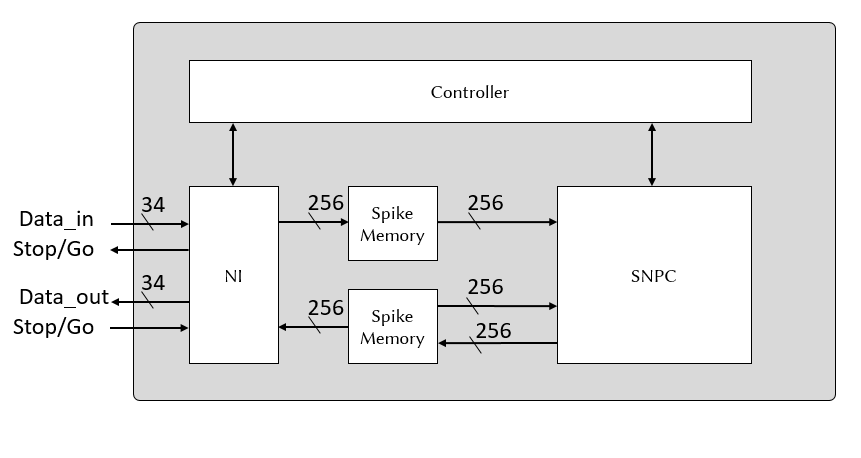}
	\caption{Processing element.}
	\label{fig:systpe}
\end{figure}

The overall architecture is shown in Figure~\ref{fig:systtop} which is based on 3D interconnect infrastructure. The inter-network interconnect is formed using routers (R) that connect inter- and intra-layers. The inter-layer interconnect is through-silicon-via (TSV) technology using model from North Carolina State University~\cite{NCSUEDA2015FreePDK3D45} for layout. 
However, the design could be applied for traditional 2D-IC or different 3D-IC technologies. 
Here, TSV-based 3D-IC is used because of its advantages on power consumption and scalability.
While the communication is handled by 3D interconnect, the computation is done by Processing Elements (PEs) where its architecture is shown in Figure~\ref{fig:systpe}. Here, the controller manages the computation by time step, where the synchronization is done via communication. The incoming spikes are stored in memory and fed to the SNPC (spiking neural processing core). The output spikes from SNPC are also stored in different memory and could be fed back to SNPC as recurrent connection. Here, the SNPC has 256 neurons (physical) and 65k synapses (learn-able) and 65k recurrent connection (fixed weight). The neuron performs using the start/end signals with the controllers to indicate the time-step (see Figure~\ref{fig:systspnc}).

\subsection{Inter-neural communication}
\begin{figure}
	\centering
	\includegraphics[width=\linewidth]{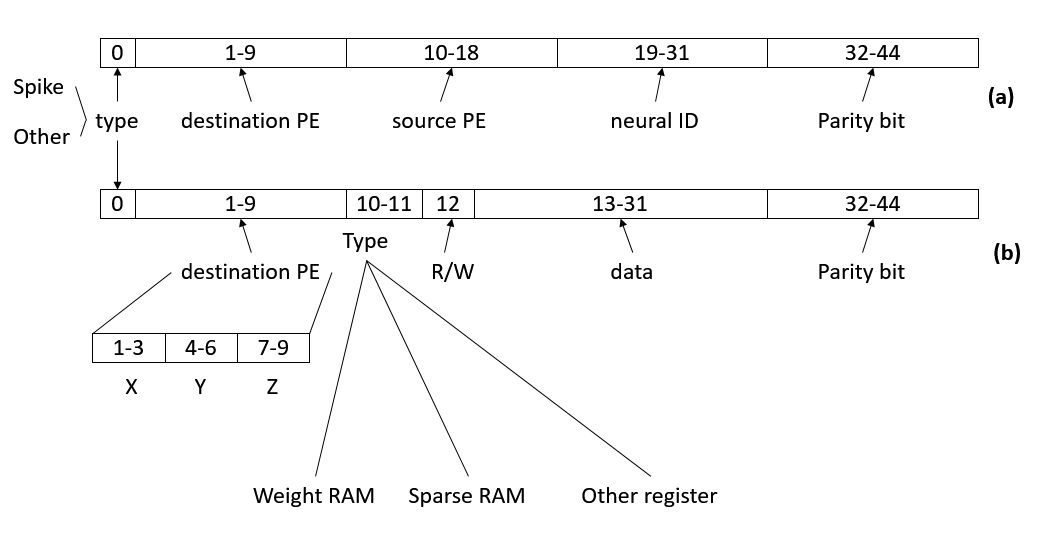}
	\caption{Flit format.}
	\label{fig:packet-format}
\end{figure}

\begin{figure}
	\centering
	\includegraphics[width=\linewidth]{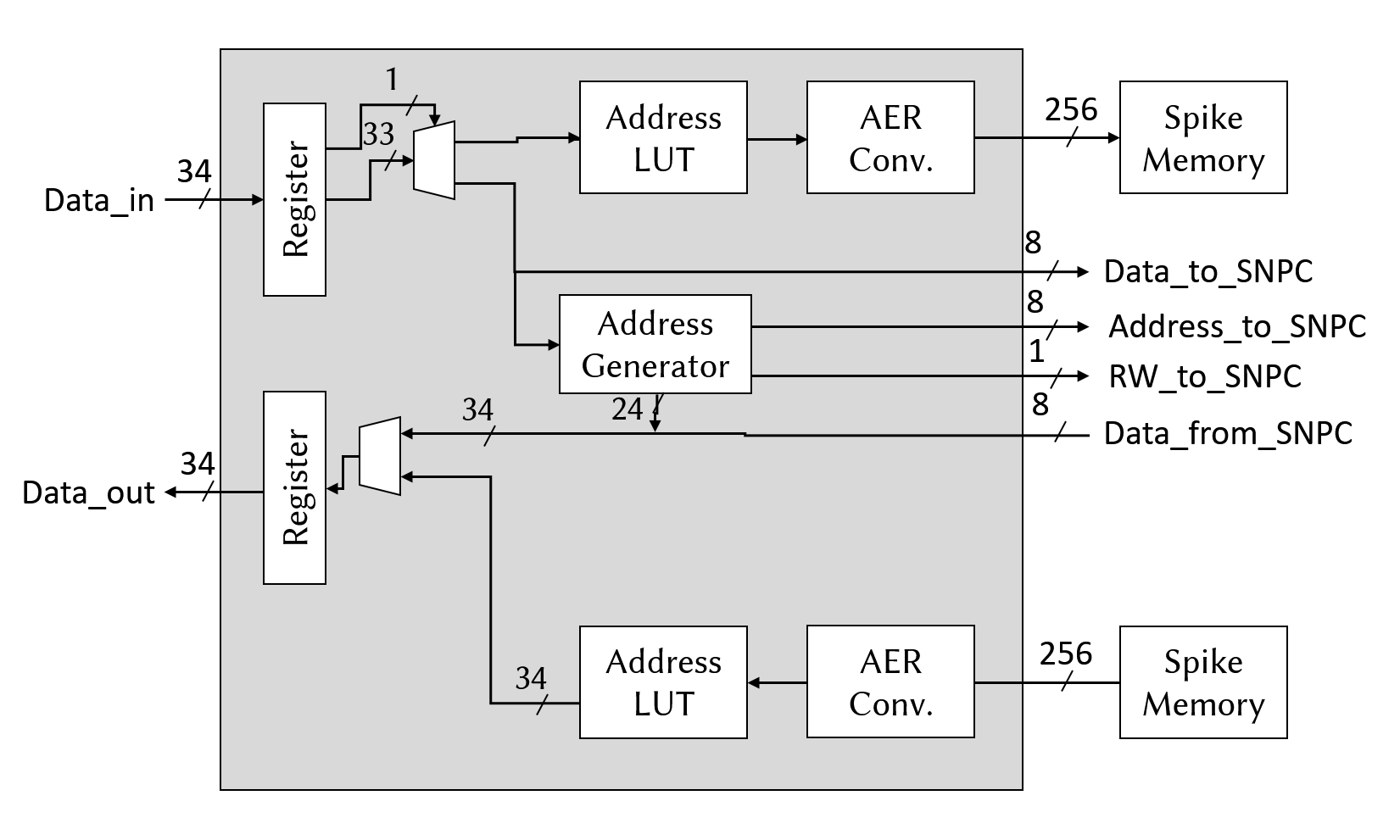}
	\caption{Network Interface Architecture.}
	\label{fig:ni}
\end{figure}

As we previously mentioned, communication is one of major challenge for hardware SNNs. To handle the communication between SNPCs, the inter-neural interconnect consists of multiple routers (R). The packet in the inter-neural is single flit as shown in Figure~\ref{fig:packet-format}. 
Here, the ``type'' field decides whether it is a spike packet or a memory access one (read/write). 
The second field is destination PE which consists of 9-bits (X-Y-Z domain) that allows maximum of 8x8x8 PEs (512 PEs). To extend the network size, adding more bits in this field is necessary. 
For spike flit, the next fields are the source PE address and the ID of the firing neural. 
For memory access flit, the next field is another type to define the accessed memory (weight, sparse or other) and is followed by data. 
Note that there is no address field since the data is read and written in burst (serially one by one). 
The last field is a parity bit to protect the integrity of a flit.
Figure~\ref{fig:ni} shows the block diagram of the Network-Interface (NI). The input spike, following the format in Figure~\ref{fig:packet-format}, is sent to the address LUT and AER converter (see Algorithm~\ref{alg:AER2Spk}), while the memory access flit is sent directly to the SNPC. 
For the memory access, there is an address generator to create write/read to the memory. 
The memory is accessed serially instead of randomly.
On the other hand, the reading memory is also initialized and the data is read from SNPC memory. 
The source address is inserted into the data field which is used to complete the flit. 
The incoming and outgoing spikes are read from the memory following the first come first serve rule.

Assuming the \textit{number of address} bits, \textit{connected PEs per PE} bits and \textit{neurons per PE} bits are $N_{address-width}$,  $N_{connected-PE}$,  $N_{neuron/PE}$, respectively. The address LUT consists of two tables: 

\begin{itemize}
\item Address to connected PE: it converts the source address of the spike to the connected PE address. 
Here, the system requires a LUT of $N_{address-width}$ banks of $N_{connected-PE}$ bits.

\item Converted the connected PE with fired neuron to the memory address in weight SRAM. Without sparsity, it requires a LUT of $N_{connected-PE}+ N_{neuron/PE}$ banks of $N_{neuron/PE} $ bits. 
	
\item With a sparsity rate $r$, there is an option of content access memory (CAM) of $N_{neuron/PE}$ banks consisting $N_{address-width}+ N_{neuron/PE}$ bits. The CAM will return the address within the memory which is the corresponding address in the weight memory.
\end{itemize}
\subsection{Input Representation}
Although the flit follows AER (Address Event Representative) protocol, the Network Interface (NI) groups them to a \textit{spike array} which represents each pre-synaptic neuron. 
Based on AER, it updates the values. Algorithm~\ref{alg:AER2Spk} shows the conversion between AER and \textit{spike array} where AER\_in consists of source PE and neural ID in Figure~\ref{fig:packet-format}. Note that with the non-sparsity, simple address conversion (adding based value) is possible. 
For the sparsity connections, a simple look-up-table could be also used. 
\begin{algorithm}[bhtp]
	\small
	\caption{AER conversion.}
	\label{alg:AER2Spk}
	
	\KwIn{AER\_in}
	\KwIn{new\_timestep}
	\KwOut{Spike\_in\tcp*{Input spike}} 
	
	Spike\_in = 0\;
	
	\While{new\_timestep !=0 }{
		pre-syn\_neuron =  decoding (AER\_in)\;
		Spike\_in = Spike\_in | 1 $\ll$ pre-syn\_neuron\;
	}
	
\end{algorithm}  
\subsection{Spiking neuro-processing core (SNPC)}
\begin{figure}[h]
	\centering
	\includegraphics[width=1\linewidth]{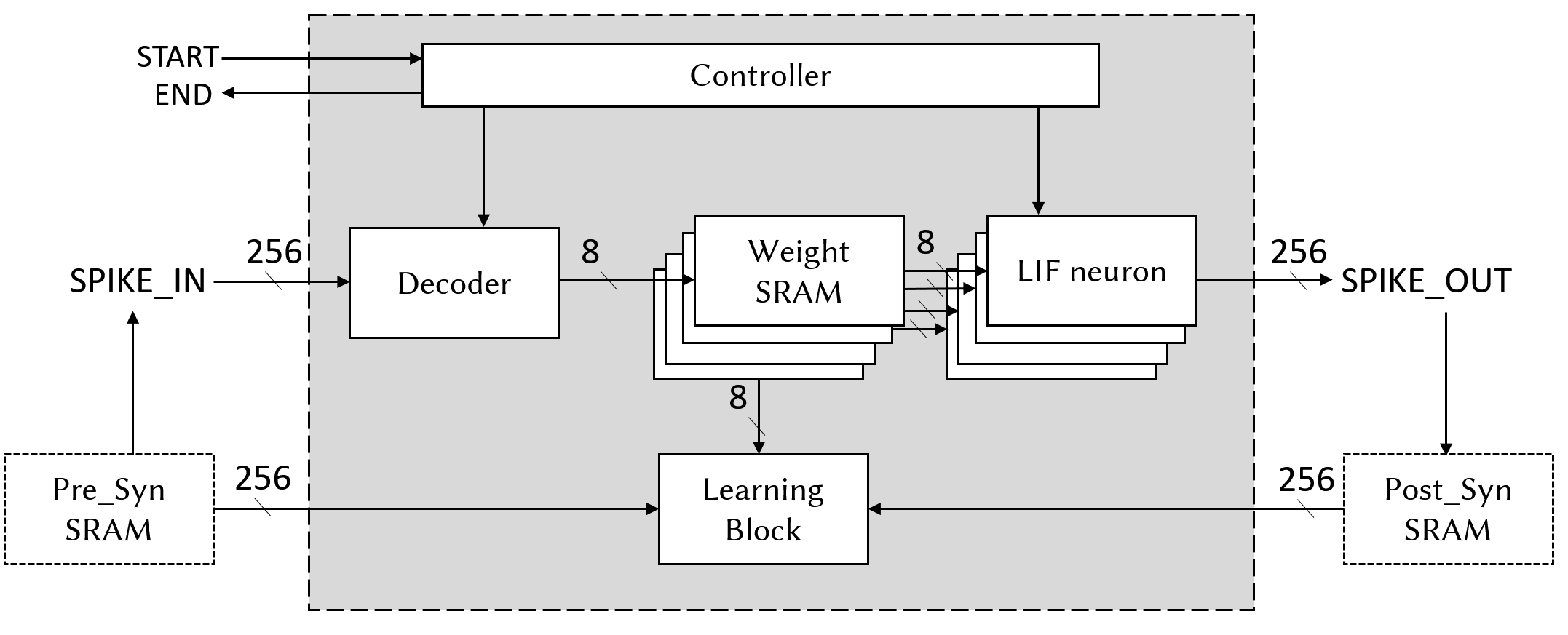}
	\caption{Spiking neural processing core.}
	\label{fig:systspnc}
\end{figure}

Figure~\ref{fig:systspnc} shows the architecture of spiking neural network. 
The input AER is decoded and grouped in the NI and the input conversion is performed  using Algorithm~\ref{alg:AER2Spk}. 
After receiving the spike array (256-bit) for each time-step, the decoder extracts the address of spike array. For instance, if there are 4 pre-synaptic neurons and the spike array is ``1010'', the decoder sends two addresses in two cycles: 1 and 3 which is corresponding to bit `1' in the spike vector. 
By generating the address, it feeds to the weight SRAM to read the weight.  
Here, it simply performs the multiplication between input spike and weight.
\\
A series of weighted input without zero will be sent to a LIF neuron, which accumulates the value, subtracts the leak and check the firing condition. 
The output spike is stored in SRAM for learning purpose. 

\subsection{Sparsity of connection}

In~\cite{Chen20184096}, the authors uses sparsity connection (i.e 50\%) method. 
Assume that we have X truth pre-synaptic neurons, $m$ neurons and we only use $n$ of pre-synaptic neurons' connection due to sparsity. 
That being said, we need a LUT of $X\times n$ bits and reduce $(X-n)mw$ bits where $w$ is bit-width of a weight. The sparsity ratio is $s= n/X$. If $s=0.5$, we need $2n^2$ bits for LUT and reduces $nmw$ for connections. Therefore, the condition of saving memory footprint is: 
\[
2n^2 < nmw \text{ or  }n < mw/2
\]

With m= 256 and w= 8, we have the saving condition is $n < 1024$.

The saving ration is:

\[
\frac{(X-n)mw - Xn}{Xmw} = 1- \frac{(mw+X)s}{mw}
\]

With m = 256, X = 786, w = 8, saving ratio is: $1-1.383s$. Therefore, $s < 0.723$ is the condition of saving memory footprint. By following the discussed conditions, designers could calculate their proper value of sparsity. We also want to note there is a trade of between sparsity and accuracy.
 
\subsection{Crossbar}

Each word of SRAM stores a weight value. In this design, 256 physical neurons requires 256 separated SRAMs. However, since the input (pre-synaptic neuron spike) is shared, these SRAMs share the reading address. Therefore, we could merge these SRAMs to reduce the area cost. 
For instance, merging 8 8-bit weights into 64-bit word SRAM, the system only needs 32 SRAMs. 
Depending on the SRAM technology and the size of the weight, designers could design the desired SRAM structure.
\begin{figure}
	\centering
	\includegraphics[width=\linewidth]{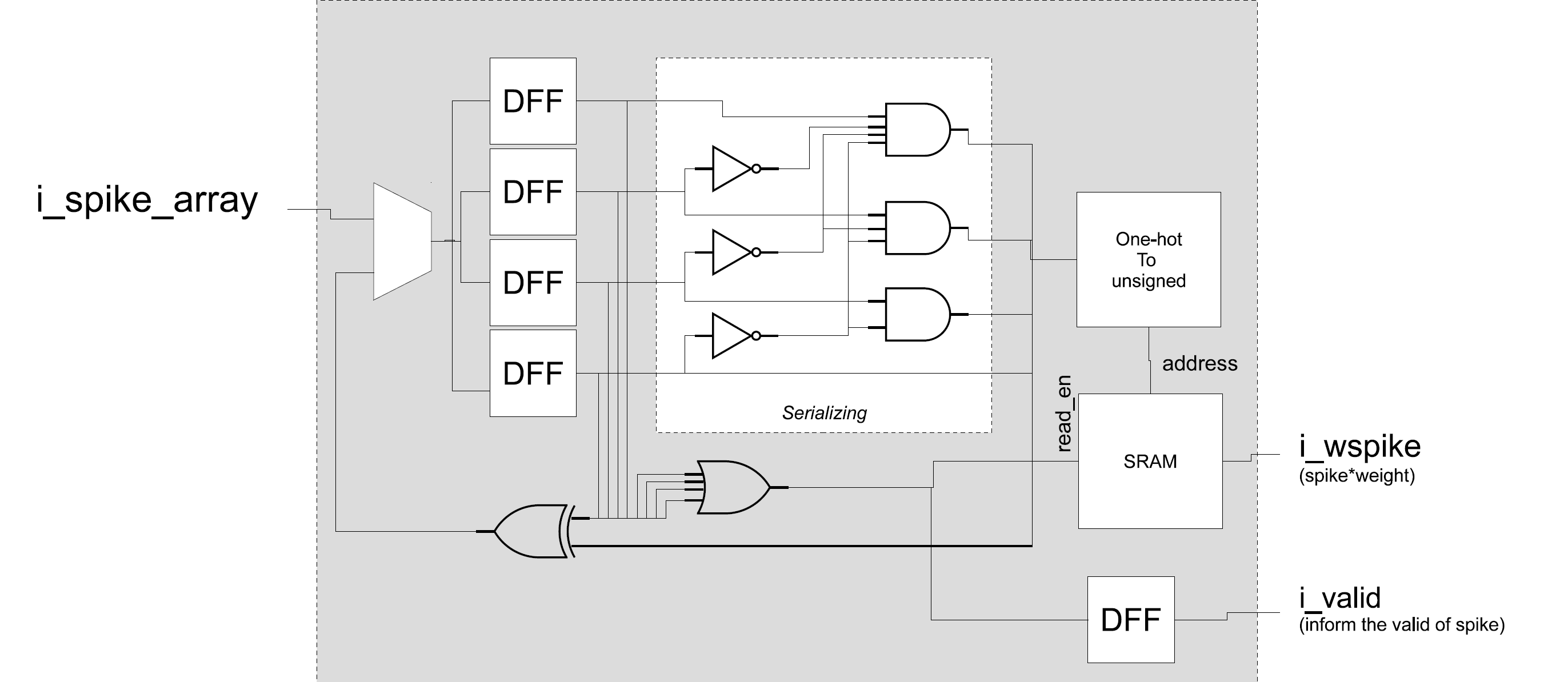}
	\caption{Decoder architecture}
	\label{fig:crossbardec}
\end{figure}
To extract the address from the decode stage, we use the architecture in Figure~\ref{fig:crossbardec}. It extracts one-hot value with least index from the spike array and converts it to index value to feed to the SRAM. 
Then, one-hot value is XORed with the spike array value to erase the one-hot bit which is feed.
The decoder method is also reused in AER conversion. Therefore, the output spike is fed to decoder to extract each spiking index. Then, the spiking indexes are converted to AER.

Using the decoder in incoming spikes could save the area cost if the number of spike is very large. For instance, with 8-bit address for 256 bit spike array, this method could save memory footprint when there are more than (256/8) 32 spikes withing a time step. 
We would like to note that with fully connected SNN, if the previous layer has a huge number of neurons, storing spike array could avoid overflow since it is too difficult to have enough memory. 
For example, for 1000 pre-synaptic neurons, we need 1000x10 bit memory to avoid overflow and 1000x10x2 for pipe-lining. 
Meanwhile, the spike array only uses 1000 and 1000x2 for these cases. 
Table~\ref{tab:comp-aer} compares the memory footprint using AER and spike array. Apparently, keeping spike array scales much better than storing AER inputs.
\begin{table}[h]
	\centering
	\caption{Comparison between keeping AER and using spike array.}
	\label{tab:comp-aer}
	\begin{tabular}{lcc}
		\hline
		\textbf{Paramter} & \textbf{AER} & \textbf{Spike array}   \\\hline
		\# pre-synaptic neuron & n  & n \\ 
		one value width & $log_2(n)$  & n \\ 
		X values width & $X\times log_2(n)$  & n \\ 
		Overflow condition & X=n & unneeded \\ 
		Overflow +pipeline & X=2n & unneeded \\ \hline
	\end{tabular}
\end{table}
\subsection{LIF neuron}
Most hardware  friendly neural architecture focuses on either Leaky-Integrate-and-Fire (LIF) or Integrate-and-Fire (IF) model due to their simplicity. 
By lowering the area cost of neuron design, more neurons could be integrated.

Theoretically, a LIF/IF neuron computation is shown in the equation  bellow:

\begin{equation}
V_j(t) = V_j(t-1) + \sum_i w_{i,j}\times x_i(t-1) -\lambda
\end{equation}

Where: 
\begin{itemize}
	\item $V_j(t)$ is the membrane potential of neuron $j$ at time step $t$,
	\item $w_{i,j}$ is the connection weight (synapse strength) between the pre-synapse neuron $i$ and the post-synapse neuron $j$.
	\item $x_i(t-1)$ is the output spike of pre-synapse neuron $i$ 
	\item $\lambda$ is the constant leaky ($\lambda=0$ for IF).
\end{itemize}

The output spike of neuron $j$ follows this equation:
\begin{equation}
x_j(t)= 
\begin{cases}
1,& \text{if } V_j(t)\geq Vt\\
0,              & \text{otherwise}
\end{cases}
\end{equation}
\begin{figure}
	\centering
	\includegraphics[width=\linewidth]{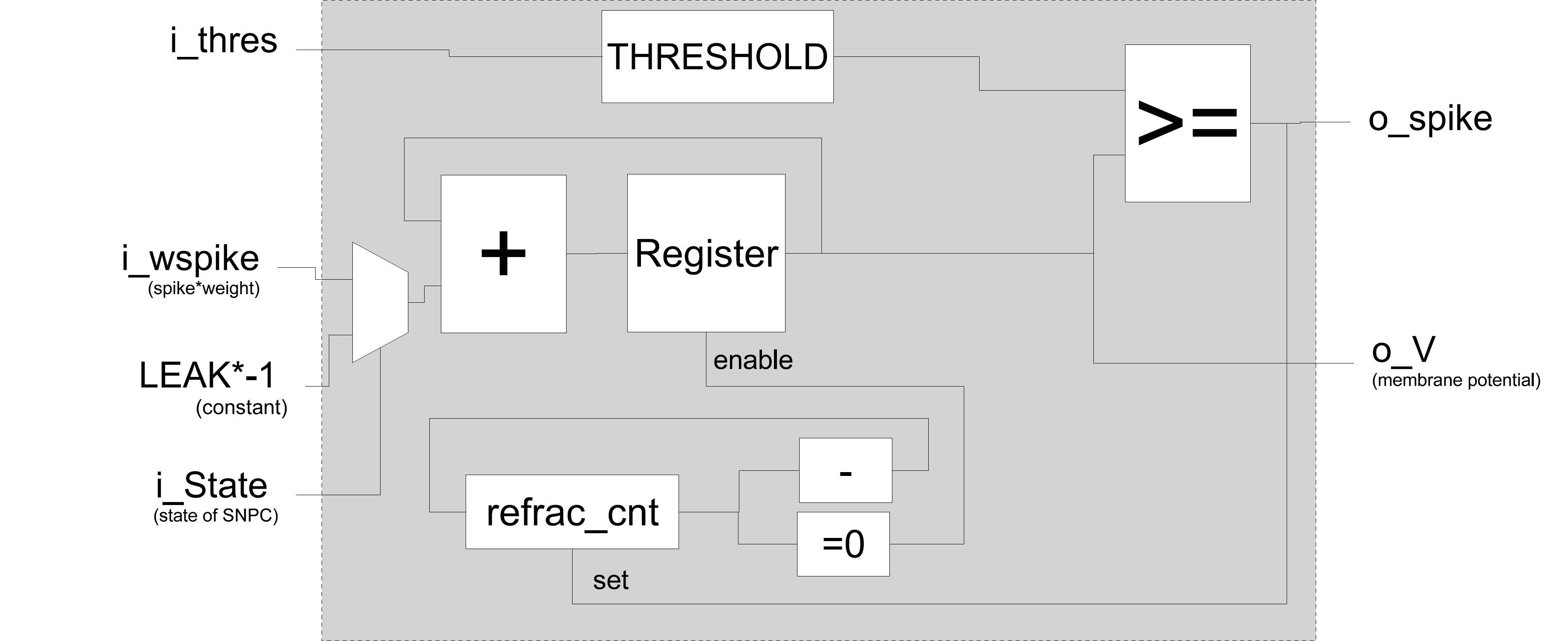}
	\caption{LIF neuron architecture}
	\label{fig:lifneuron}
\end{figure}

Figure~\ref{fig:lifneuron} shows the architecture of LIF neuron. The weighted input (i\_wspike) is fed into an adder+register structure to accumulate the value. At the end of time step, the inverted value of leak is fed to reduce the membrane potential. The membrane potential is compared with threshold to check the firing condition. If the neuron fire, it sets the count down refac\_cnt to keep the neuron stops working for several time step.
\subsection{Recurrent connection} 

Beside the crossbar for forwarding connection, there is an recurrent connection in our design. Note that the ANN-SNN conversion does not need this step. 
The architecture is similar to Figure~\ref{fig:crossbardec}, where the output spike array is converted to a series of spiking index. 
These indexes are sent to the the PE of the same layer to perform the recurrent. Also, a fixed and negative weight crossbar is used without a RAM because the weight is fixed. Therefore, we could reduces the area cost of recurrent connection. 

For inter-neuron communication, the recurrent is detected by checking whether a source neuron belongs to the same layer with current one. 
\subsection{STDP Learning Block}
 The learning block of SNPC follows the STDP learning rule:
 \begin{itemize}
 	\item It first stores the pre and post-synaptic neuron spike.
 	\item Once there is a post-synaptic neuron spike, it increases the weights of previously fired pre-synaptic neurons and reduces the weights of lately fired pre-synaptic neurons.
 \end{itemize}
 \begin{figure}
 	\centering
 	\includegraphics[width=\linewidth]{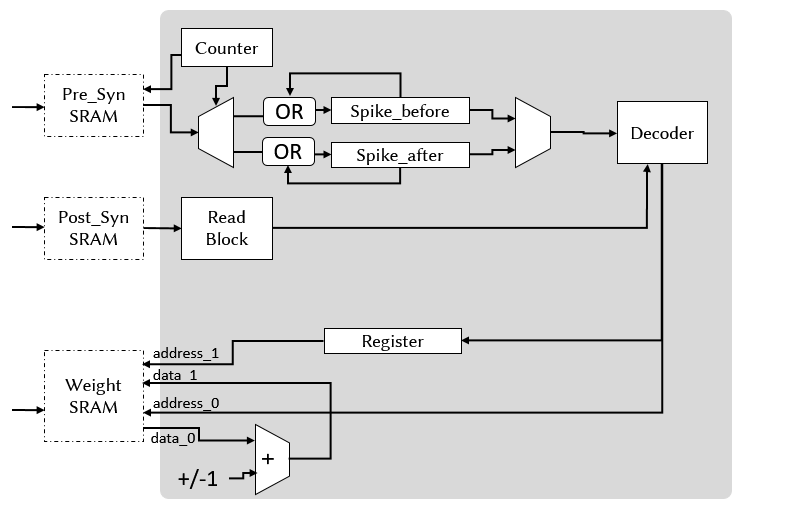}
 	\caption{Learning Block.}
 	\label{fig:lb}
 \end{figure}
 
Figure~\ref{fig:lb} shows the learning block architecture. A
First, the post\_syn RAM (post-synaptic neuron spike SRAM) is read to check whether the calculated time step has outgoing spike. If not, it skips the calculation. 
Then, it reads the pre-synaptic spike from pre\_syn SRAM to find the pre-synaptic neuron fires before and after the calculating time step. 
We would like to note that the calculating time step is performed after the time step of the LIF core since we need to know the spike after it. 
The read value is OR with value stored in a register. 
The final results are two vectors representing the pre-synaptic neurons.
 
The decoder read these values using the same architecture of the SNPC decoder, then it extracts the addresses one by one, feeds to address\_0 and obtains the weight at data\_0 at the following cycle. 
Depending on the ''before/after'' registers, it increase/decrease by 1 and write back to the weight SRAM.
Our STDP algorithm also follows two main rules: (1) \textit{adaptive threshold:} during learning, the threshold of a neuron is increased once it fires and slowly reduced; (2) weight normalization: the sum of all weights of a neuron is constant.
\subsection{Software platform}
In this study, we present two hardware platforms. The first one is based on the work in \textit{Diehl et al.}~\cite{Diehl2015Fast} which is a direct conversion of ANN to SNN. The second one is based on a modified version of BindsNet - a Python-based SNN simulator~\cite{Hazan2018BindsNET}. We implemented both derived SNN systems in hardware (described later). 
\subsubsection{ANN to SNN Conversion}

To perform the ANN to SNN conversion, we used the MATLAB DeepLearnToolbox~\cite{Palm2012Prediction} which provides both convolutional and full-connected networks. This part of code helps training normal ANNs for implementaiton. 
The conversion to SNNs with normalization is done by ~\cite{Diehl2015Fast}. 
His conversion targets MNIST benchmarks; however, a similar platform of conversion is found in~\cite{Rueckauer2017Conversion}.

For hardware implementation, the weights, thresholds for this conversion are needed. This SNN version is based on IF (integrate and fire) spiking neuron model. Thanks to the normalization by ~\cite{Diehl2015Fast}, the threshold is fixed to 1. 
Furthermore, we also used fixed-point for the weights. For example, if we use 7 fractional bits, the weight and threshold is multiplied to $2^8$. The simulation is performed with fixed-bit to observe the accuracy. At the end, we conclude that 7-bit is the most suitable one. Further results could be seen in the evaluation section. 
We have to note that by directly converting ANN to SNN, we could explore an enormous amount of existing algorithms and methods of ANN. Therefore, training offline using ANN and convert to fixed-point values to execute in hardware is a possible solution.
\subsubsection{Spiking Neural Network simulator}
Although ANN to SNN conversion is a helpful method in re-implementing ANN algorithms, it is not a natural approache. To solve this problem, we used here a spiking neural network simulator call BindsNet~\cite{Hazan2018BindsNET} to help enhancing the design. 

BindsNet is built on the famous \textit{Pytorch}~\cite{Paszke2017Pytorch} deep learning neural network library to implement the function of neural network. Although there are some existing models for SNN within BinsNet, their target is to simulate the function of brain, not the hardware design. To do so, we implement our own Python package based on BindsNet to reuse their functions. On top of that, we build our own hardware friendly SNN system. Since BindsNet already implements the spiking generator and several other modules, we reused them in our own package. Here, we provides three main features:

\begin{itemize}
	\item First, we provide a simple LIF core and SNN as we previously designed in Section~\ref{sec:arch}. The Python snippet could be found bellow. Note that we use adaptive threshold here, which is fixed during interference.
	
	\item Our own simplified STDP version: fixed change (Fixed $\Delta w$ ) instead of time-dependent change (Fixed $\Delta w$ ) is applied into hardware. 
	
	\item Our off-line training flow. The weights and threshold are converted to fixed bits instead of floating points. These values are also loaded to our software SNN to evaluate the accuracy. They are also exported to binary files which are later read by hardware as off-line training.
	
\end{itemize}

Note that the original STDP rule follows the spike traces:

\[
\Delta w = \begin{cases}
\eta_{post}\chi_{pre} \text{ on post-synaptic spike;}\\
- \eta_{pre}\chi_{post} \text{ on pre-synaptic spike;}
\end{cases}
\]

where the spike traces ($\chi_{pre}$  and $\chi_{post}$) are set to 1 at the events then decaying to 0. Here, we call this  \textit{Adaptive $\Delta w$} while our method is called  \textit{Fixed $\Delta w$}.

In conclusion, we built the software model for SNNs that approximate the accuracy of hardware model. We actually could further apply fixed point conversions for better accuracy. The same method could be re-used from the Matlab model.

\subsection{Integration of SNN into On-chip Communication} \label{sec:inte}
\begin{figure}
	\centering
	\includegraphics[width=\linewidth]{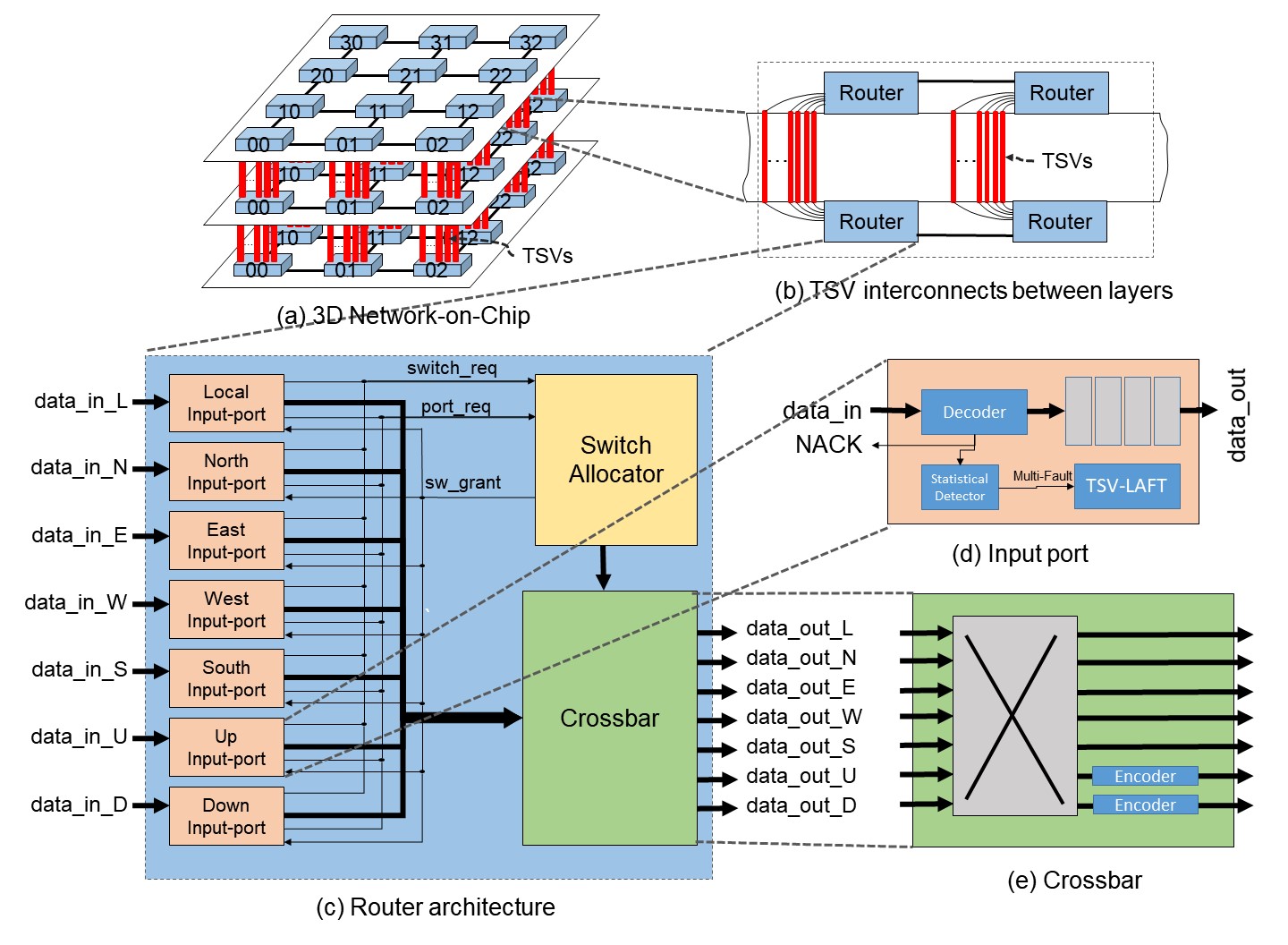}
	\caption{On-chip communication architecture.}
	\label{fig:figarchnoctorouter}
\end{figure}

In the last section, we have shown the overall architecture and the design in hardware and software for SNN. In this section, we discuss the integration of the SNN to the on-chip communication. As already shown in Figure~\ref{fig:systtop}, the overall design consists of PE (NI and SNPC) and the routing block. 
Figure~\ref{fig:figarchnoctorouter} shows more detail of the on-chip communication. The 3D-mesh topology is formed to support the 3D-ICs and the SNNs  where the interconnect between layers are based on TSVs. Each router has seven ports for seven directions. Here the incoming flit is routed based on the destination field the flit structure in Figure~\ref{fig:packet-format}. 
The routed flit is sent to a proper port via a crossbar which is composed of multiplexers and demultiplexers. Thanks to the routing ability of the router, a flit can travel from one node to any node within the system.

%
%
Since the NoC is fault sensitive, we have developed several fault-tolerance methods for correcting hard faults in buffer\cite{Ahmed2016Adaptive}, crossbar\cite{Ahmed2016Adaptive},  interconnect (intra-\cite{Ahmed2016Adaptive} and inter-\cite{Dang2017Scalable} layer), and soft-errors~\cite{Dang2018Parity,Dang2015Soft}. All of these techniques are well integrated in the on-chip communication to help it recover from both permanent and transient fault.

\section{Evaluation} \label{sec:eva}
This section first provides the evaluation method for our SNNs. Then, it shows the hardware results in ASIC 45nm technology. In the following part, we evaluate the SNNs under the popular MNIST benchmarks with two SNN model: ANN to SNN conversion and unsupervised STDP.
\subsection{Evaluation methodology}
We first perform the software platform simulation to have a preliminary result for the design. Then, the hardware implementation is performed.
For evaluations, we select MNIST~\cite{LeCun1998MNIST} which is one of most popular dataset. This is also the current limitation for SNN learning algorithm. 
The data (pixel) is normalized with the maximum value (256) and encoded using Poisson distribution.
\subsection{ANN to SNN conversion for multi-layer networks}
\begin{figure}[htb!]
	\centering
	\includegraphics[width=1\linewidth]{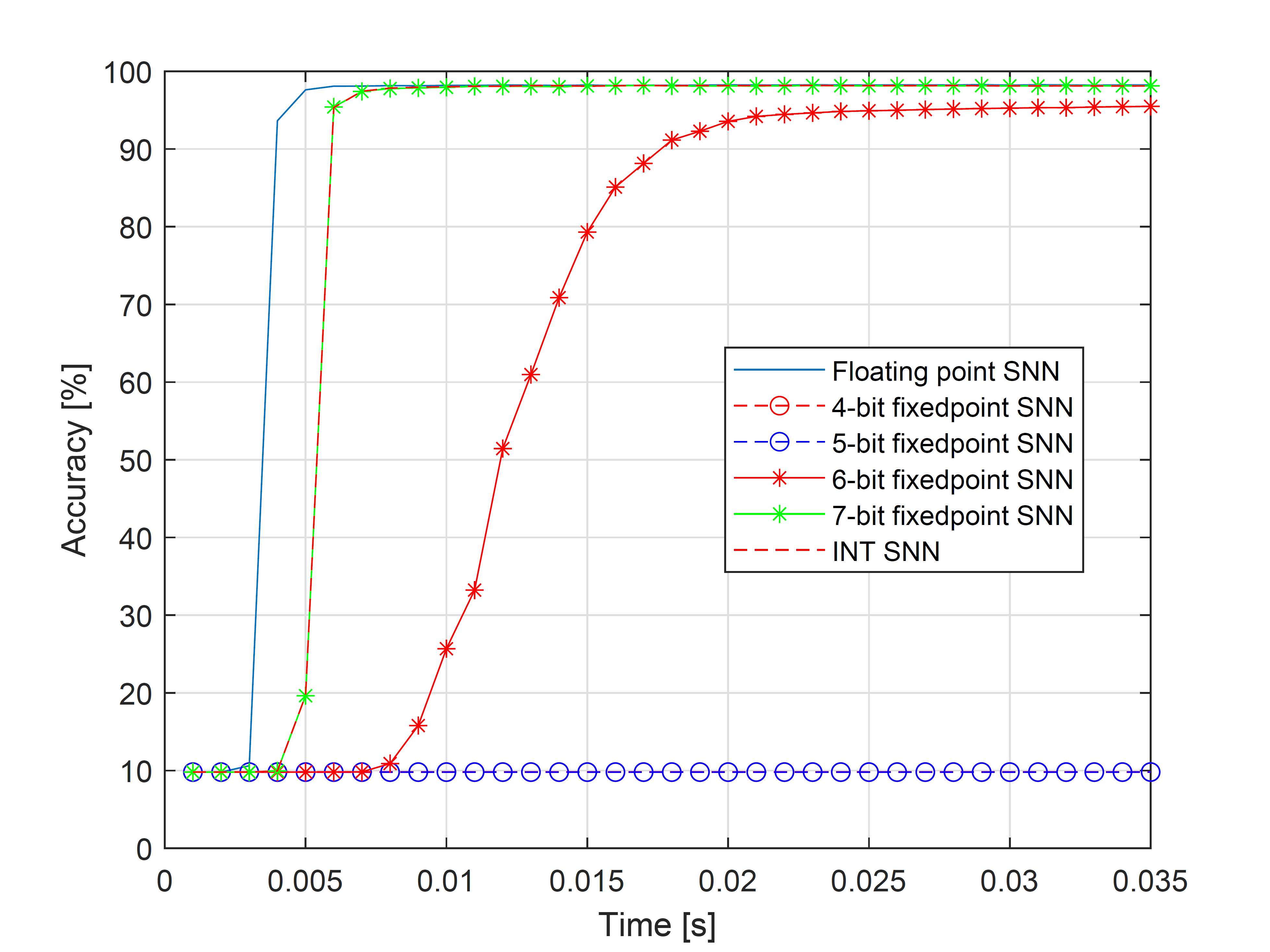}
	\caption{Accuracy result of ANN-to-SNN conversion (784:48:10) for MNIST. Time step is 0.001 second. }
	\label{fig:small-ann}
\end{figure}

\begin{figure}[htb!]
	\centering
	\includegraphics[width=1\linewidth]{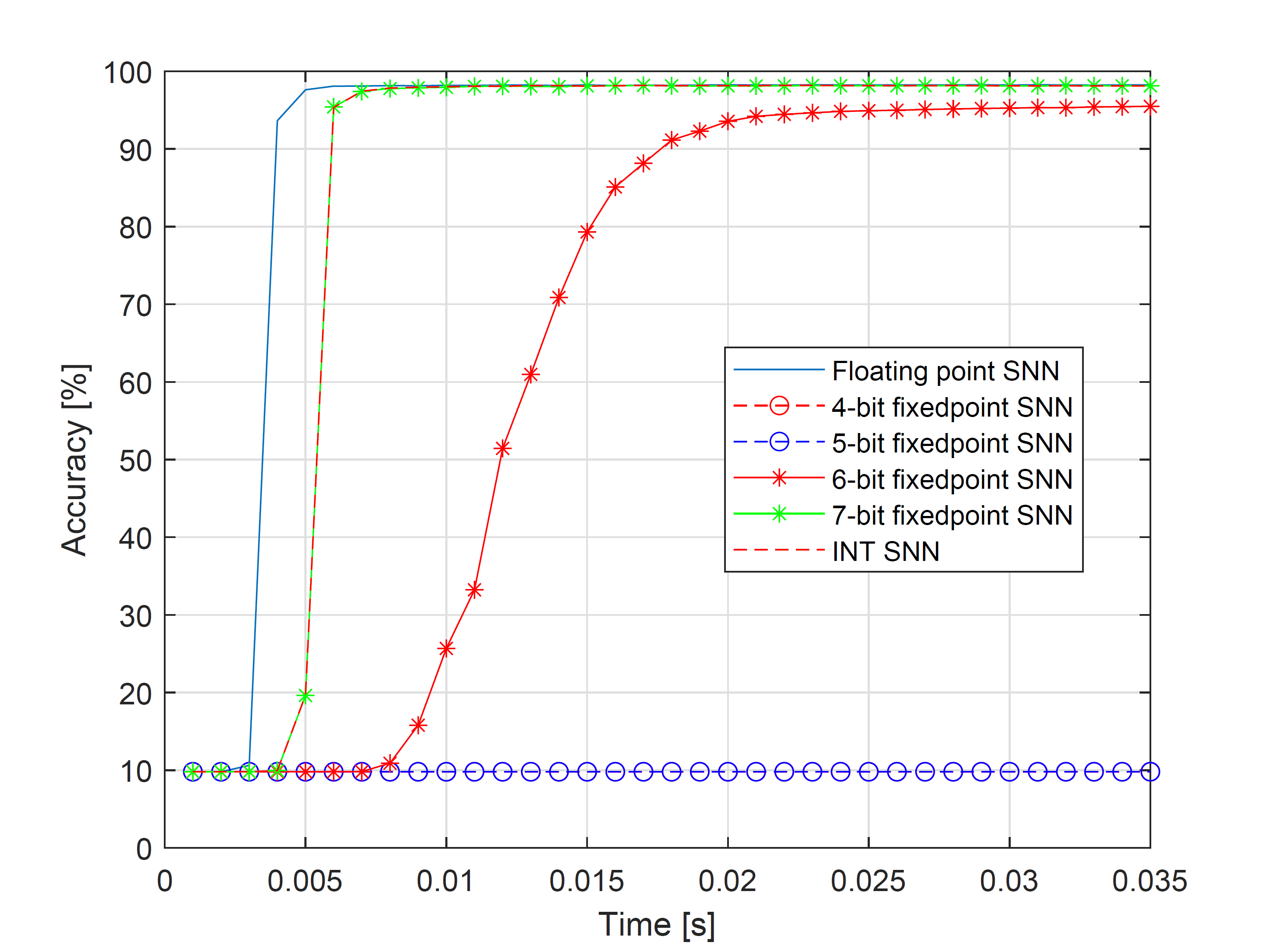}
	\caption{Accuracy result of ANN-to-SNN conversion (784:1200:1200:10) for MNIST. Time step is 0.001 second.}
	\label{fig:full-ann}
\end{figure}

For the ANN to SNN conversion, we use two networks: 784:1200:1200:10 and 784:48:10 for MNIST dataset. 
Figure~\ref{fig:small-ann} and~\ref{fig:full-ann} show the 784:48:10 and 784:1200:1200:10 SNNs, respectively. A time step is 0.0001 second which make the total number of time steps 350. Here, we also evaluate the fixed point SNN where we keep the least significant bit in representation. 
We also evaluated INT SNN which converts 7-bit fixed point to all integer which give the result for hardware implementation. 
We can easily see the drop in terms of accuracy when comparing the floating point SNN and the fixed point ones. The drops are significant when the number of representing bit is less than 5 in 784:48:10  and 6 in 784:1200:1200:10. The main reason is the larger and deeper network will accumulate the differences in values which make more inaccurate results. Nevertheless, we can easily see that 7-bit fixed point is the best candidate for implementation which provides nearly identical accuracy at the end and only slower response time. 
For the smaller network, 6- and 5-bit fixed point versions are considerable; however, the final results are bellow 95\% which is not high in the state-of-the-art standard. For the large network (784:1200:1200:10), we can see the 7-bit or INT SNN reach the floating point SNNs. The floating point, 7-bit and INT SNN have the accuracy of 98.25\%, 98.13\% and 98.12\%, respectively.

On the other hand, the larger network of floating point SNN also saturates faster at the 45th time step where the 7-bit and INT ones stature around the 55th. If the system cuts the operation at this point, it could saves nearly 84.28\% of computation time. By using the clock gating~\cite{Mahmoodi2008Ultra} where the energy could be saved around 69\% at zero data switching activity, the power consumption could be dramatically dropped. More power could be reserved with power gating technique.

\subsection{Unsupervised STDP}

The previous section evaluates the ANN to SNN conversion. In this section, we evaluate the unsupervised STDP method. Here we adopt the network of Diehl \& Cook~\cite{Diehl2015Unsupervised} whith a sligt modification. Instead of using an inhibitory layer; we use recurrent connection which reduces the network into half size for hardware implementation. The recurrent version could be found in the work \textit{Hazel et al.}~\cite{Hazan2018BindsNET}. Furthermore, we simplify the architecture to be identical to hardware implementation. 
The network size of Diehl \& Cook is 784:N:N while our network is 784:N. 

\begin{table}[]
	\centering
	\scriptsize 
	\caption{Accuracy result of STDP learning for SNNs.}
	\label{tab:stdp-snn}
	\begin{tabular}{lllll}
		\hline
		\textbf{N}   &  \textbf{Diehl \& Cook}~\cite{Diehl2015Unsupervised} & \textbf{Adaptive $\Delta w$ }    & \textbf{Fixed $\Delta w$} & \textbf{HW SNN} \\ \hline
		100 & 79.44\%        & 79.31\% & 71.08\%                                                           & 71.32\%          \\ 
		400 & 88.87 \%       & 84.94\% & 83.93\%                                                           & 84.05\%          \\ \hline
	\end{tabular}
\end{table}

Table~\ref{tab:stdp-snn} shows the accuracy comparison between Diehl \& Cook~\cite{Diehl2015Unsupervised} network and our three versions: (1) Adaptive $\Delta w$:  weight-dependent change ($\Delta w = w \times \text{learning\_rate}$); (2)  Fixed $\Delta w$: constant weight change; and (3) HW SNN: constant weight change and 8-bit fixed point for all parameters (weight, threshold).

Comparing between the adaptive $\Delta w$ and Diehl \& Cook~\cite{Diehl2015Unsupervised}, we could observe a small drop in accuracy by using our simple RTL-like model. The drop is insignificant in N=100 but around 3.8\% in N=400. However, by completely convert to RTL model, the drop become more significant. The accuracy loss is 8.12\% and 2.82\% for N=100 and 400, respectively. While the change between HW SNN and fixed  $\Delta w$  is not significant, we easily observe the changing $\Delta w$ affects the accuracy.

\subsection{Spiking computation module}
Table~\ref{tab:comp} shows the comparison between our and existing works. 
Note that due to lack of library, we use register files instead of proper memory which makes the area cost much higher than SRAM design.

Comparing between the  existing work, then proposed our has lower area cost among them. When comparing with the work in \textit{Seo et al.}~\cite{Seo201145nm}, our system's area cost is smaller despite of having 256 physical neurons, 8-bit instead of 1-bit and register instead of SRAM. The work by \textit{Frenkel et al.}\cite{Frenkel20180.086}, \textit{Akyopyan et al.}\cite{Akopyan2015TrueNorth} and  is also smaller, but by converting to 45 $nm$ and 8 bit weight, our design is still smaller. 

\begin{table*}[h]
	\centering
	\scriptsize
	\caption{Comparison of current work with existing works.}
	\label{tab:comp}
	\begin{tabular}{|l|l|l|l|l|l|l|l|}
		\hline
		\textbf{Author}                         & \textit{Benjanmin et al.}\cite{Benjamin2014Neurogrid} & \textit{Painkras et al.}\cite{Painkras2013SpiNNaker}  & \textit{Frenkel et al.}\cite{Frenkel20180.086} & \textit{Seo et al.}~\cite{Seo201145nm}  & \textit{Akyopyan et al.}\cite{Akopyan2015TrueNorth} & \textit{David et al.}~\cite{Davies2018Loihi}      & \textbf{Ours}          \\ \hline
		\textbf{Publication}                    & PIEEE, 2014      & JSSC, 2014       & TBioCAS, 2019  & CICC, 2011  & TCAD, 2015      & IEEE Micro, 2018  & This work     \\ \hline
		\textbf{Implementation}                 & Analog           & Digital/Software & Digital        & Digital     & Digital         & Digital           & Digital       \\ \hline
		\textbf{Technology}                     & 0.18 $\mu m$     & 0.13 $\mu m$     & 28 $nm$ FDSOI  & 45 $nm$ SOI & 28 $nm$         & 14 $nm$ FinFET    & 45 $nm$       \\ \hline
		\textbf{Core area }$mm^2$ & 168              & 3.75             & 0.086          & 0.8         & 0.095           & 0.4               &    0.205608           \\ \hline
		\textbf{Weight storage}        & Off-chip         & Off-chip         & 4-bit SRAM     & 1-bit       & 1-bit SRAM      & 1- to 9- bit SRAM & 8-bit register file    \\ \hline
		\textbf{Learning}                       & Offline          & Programmable     & SDSP           & Prob. STDP  & Offline         & Programmable      & Offline, STDP \\ \hline
		\textbf{Neurons per core}               & 64k              & max. 1000        & 256            & 256         & 256             & max. 1024         & 256           \\ \hline
		\textbf{Synapses per core}              & -                & -                & 64k            & 64k          & 64k             & 1M to 114k        & 64k           \\ \hline
	\end{tabular}
\end{table*}

The Loihi chip by Intel~\cite{Davies2018Loihi} is the largest design; however, they have much higher number of neurons and synapses as well as embeds programmable learning algorithm.
\section{Conclusion} \label{sec:concl}
This work presents a design and implementation for Spiking Neural Network in hardware and software. This work is a promising step for further study the implementation and advanced optimization of hardware SNNs. Under ANN to SNN conversion, our SNN could reach 99\% accuracy. The pure SNN approach, such as STDP, has lower accuracy; however, it is still promising result which could be further optimized. 
Furthermore, we also present an efficient  method to localize up to seven permanent faults in on-chip communication and remove transient faults. 

Further study in learning algorithm, memory technology could help advance the SNN design. Further investigation is needed to study the behavior of faults for a robust system architecture. 



\begin{thebibliography}{10}
	\providecommand{\url}[1]{#1}
	\csname url@samestyle\endcsname
	\providecommand{\newblock}{\relax}
	\providecommand{\bibinfo}[2]{#2}
	\providecommand{\BIBentrySTDinterwordspacing}{\spaceskip=0pt\relax}
	\providecommand{\BIBentryALTinterwordstretchfactor}{4}
	\providecommand{\BIBentryALTinterwordspacing}{\spaceskip=\fontdimen2\font plus
		\BIBentryALTinterwordstretchfactor\fontdimen3\font minus
		\fontdimen4\font\relax}
	\providecommand{\BIBforeignlanguage}[2]{{%
			\expandafter\ifx\csname l@#1\endcsname\relax
			\typeout{** WARNING: IEEEtran.bst: No hyphenation pattern has been}%
			\typeout{** loaded for the language `#1'. Using the pattern for}%
			\typeout{** the default language instead.}%
			\else
			\language=\csname l@#1\endcsname
			\fi
			#2}}
	\providecommand{\BIBdecl}{\relax}
	\BIBdecl
	
	\bibitem{Mead1990Neuromorphic}
	C.~{Mead}, ``Neuromorphic electronic systems,'' \emph{Proceedings of the IEEE},
	vol.~78, no.~10, pp. 1629--1636, Oct 1990.
	
	\bibitem{Davies2018Loihi}
	M.~{Davies} \emph{et~al.}, ``Loihi: A neuromorphic manycore processor with
	on-chip learning,'' \emph{IEEE Micro}, vol.~38, no.~1, pp. 82--99, January
	2018.
	
	\bibitem{Akopyan2015TrueNorth}
	F.~{Akopyan} \emph{et~al.}, ``Truenorth: Design and tool flow of a 65 mw 1
	million neuron programmable neurosynaptic chip,'' \emph{IEEE Transactions on
		Computer-Aided Design of Integrated Circuits and Systems}, vol.~34, no.~10,
	pp. 1537--1557, Oct 2015.
	
	\bibitem{Furber2014SpiNNaker}
	S.~B. {Furber} \emph{et~al.}, ``The spinnaker project,'' \emph{Proceedings of
		the IEEE}, vol. 102, no.~5, pp. 652--665, May 2014.
	
	\bibitem{Benjamin2014Neurogrid}
	B.~V. {Benjamin} \emph{et~al.}, ``Neurogrid: A mixed-analog-digital multichip
	system for large-scale neural simulations,'' \emph{Proceedings of the IEEE},
	vol. 102, no.~5, pp. 699--716, May 2014.
	
	\bibitem{Schemmel2010wafer}
	J.~{Schemmel} \emph{et~al.}, ``A wafer-scale neuromorphic hardware system for
	large-scale neural modeling,'' in \emph{Proceedings of 2010 IEEE
		International Symposium on Circuits and Systems}, May 2010, pp. 1947--1950.
	
	\bibitem{Vu2019Comprehensive}
	T.~H. Vu \emph{et~al.}, ``Comprehensive analytic performance assessment and
	k-means based multicast routing algorithm and architecture for {3D}-{NoC} of
	spiking neurons,'' \emph{J. Emerg. Technol. Comput. Syst.}, vol.~15, no.~4,
	pp. 34:1--34:28, Oct. 2019.
	
	\bibitem{Vu2019Fault}
	T.~H. {Vu} \emph{et~al.}, ``Fault-tolerant spike routing algorithm and
	architecture for three dimensional noc-based neuromorphic systems,''
	\emph{IEEE Access}, vol.~7, pp. 90\,436--90\,452, 2019.
	
	\bibitem{Diehl2016Conversion}
	P.~U. {Diehl} \emph{et~al.}, ``Conversion of artificial recurrent neural
	networks to spiking neural networks for low-power neuromorphic hardware,'' in
	\emph{2016 IEEE International Conference on Rebooting Computing (ICRC)}, Oct
	2016, pp. 1--8.
	
	\bibitem{Kim2016Neurocube}
	D.~{Kim} \emph{et~al.}, ``Neurocube: A programmable digital neuromorphic
	architecture with high-density 3d memory,'' in \emph{2016 ACM/IEEE 43rd
		Annual International Symposium on Computer Architecture (ISCA)}, June 2016,
	pp. 380--392.
	
	\bibitem{NCSUEDA2015FreePDK3D45}
	{NCSU Electronic Design Automation}, ``{FreePDK3D45 3D-IC process design
		kit},'' \url{http://www.eda.ncsu.edu/wiki/FreePDK3D45:Contents}, (accessed
	16.06.16).
	
	\bibitem{Chen20184096}
	G.~K. Chen \emph{et~al.}, ``A 4096-neuron 1m-synapse 3.8-pj/sop spiking neural
	network with on-chip stdp learning and sparse weights in 10-nm finfet cmos,''
	\emph{IEEE Journal of Solid-State Circuits}, vol.~54, no.~4, pp. 992--1002,
	2018.
	
	\bibitem{Diehl2015Fast}
	P.~U. {Diehl} \emph{et~al.}, ``Fast-classifying, high-accuracy spiking deep
	networks through weight and threshold balancing,'' in \emph{2015
		International Joint Conference on Neural Networks (IJCNN)}, July 2015, pp.
	1--8.
	
	\bibitem{Hazan2018BindsNET}
	H.~Hazan \emph{et~al.}, ``Bindsnet: A machine learning-oriented spiking neural
	networks library in python,'' \emph{Frontiers in Neuroinformatics}, vol.~12,
	p.~89, 2018.
	
	\bibitem{Palm2012Prediction}
	R.~B. Palm, ``Prediction as a candidate for learning deep hierarchical models
	of data,'' Master's thesis, 2012.
	
	\bibitem{Rueckauer2017Conversion}
	B.~Rueckauer \emph{et~al.}, ``Conversion of continuous-valued deep networks to
	efficient event-driven networks for image classification,'' \emph{Frontiers
		in Neuroscience}, vol.~11, p. 682, 2017.
	
	\bibitem{Paszke2017Pytorch}
	A.~Paszke \emph{et~al.}, ``Pytorch: Tensors and dynamic neural networks in
	python with strong gpu acceleration,'' \emph{PyTorch: Tensors and dynamic
		neural networks in Python with strong GPU acceleration}, vol.~6, 2017.
	
	\bibitem{Ahmed2016Adaptive}
	A.~B. Ahmed and A.~B. Abdallah, ``Adaptive fault-tolerant architecture and
	routing algorithm for reliable many-core 3d-noc systems,'' \emph{Journal of
		Parallel and Distributed Computing}, vol. 93-94, pp. 30 -- 43, 2016.
	
	\bibitem{Dang2017Scalable}
	K.~N. Dang \emph{et~al.}, ``Scalable design methodology and online algorithm
	for {TSV}-cluster defects recovery in highly reliable {3D-NoC} systems,''
	\emph{IEEE Transactions on Emerging Topics in Computing}, in press.
	
	\bibitem{Dang2018Parity}
	K.~Dang and X.-T. Tran, ``Parity-based ecc and mechanism for detecting and
	correcting soft errors in on-chip communication,'' in \emph{2018 IEEE 12th
		International Symposium on Embedded Multicore/Many-core Systems-on-Chip
		(MCSoC)}.\hskip 1em plus 0.5em minus 0.4em\relax IEEE, 2018, pp. 154--161.
	
	\bibitem{Dang2015Soft}
	K.~N. Dang \emph{et~al.}, ``Soft-error resilient 3d network-on-chip router,''
	in \emph{2015 IEEE 7th International Conference on Awareness Science and
		Technology (iCAST)}.\hskip 1em plus 0.5em minus 0.4em\relax IEEE, 2015, pp.
	84--90.
	
	\bibitem{LeCun1998MNIST}
	Y.~LeCun, ``The mnist database of handwritten digits,'' \emph{http://yann.
		lecun. com/exdb/mnist/}, 1998.
	
	\bibitem{Mahmoodi2008Ultra}
	H.~Mahmoodi \emph{et~al.}, ``Ultra low-power clocking scheme using energy
	recovery and clock gating,'' \emph{IEEE transactions on very large scale
		integration (VLSI) systems}, vol.~17, no.~1, pp. 33--44, 2008.
	
	\bibitem{Diehl2015Unsupervised}
	P.~U. Diehl and M.~Cook, ``Unsupervised learning of digit recognition using
	spike-timing-dependent plasticity,'' \emph{Frontiers in computational
		neuroscience}, vol.~9, p.~99, 2015.
	
	\bibitem{Seo201145nm}
	J.~{Seo} \emph{et~al.}, ``A 45nm cmos neuromorphic chip with a scalable
	architecture for learning in networks of spiking neurons,'' in \emph{2011
		IEEE Custom Integrated Circuits Conference (CICC)}, Sep. 2011, pp. 1--4.
	
	\bibitem{Frenkel20180.086}
	C.~Frenkel \emph{et~al.}, ``{A 0.086-mm$^2$ 12.7-pj/sop 64k-synapse 256-neuron
		online-learning digital spiking neuromorphic processor in 28-nm CMOS},''
	\emph{IEEE transactions on biomedical circuits and systems}, vol.~13, no.~1,
	pp. 145--158, 2018.
	
	\bibitem{Painkras2013SpiNNaker}
	E.~Painkras \emph{et~al.}, ``Spinnaker: A 1-w 18-core system-on-chip for
	massively-parallel neural network simulation,'' \emph{IEEE Journal of
		Solid-State Circuits}, vol.~48, no.~8, pp. 1943--1953, 2013.
	
\end{thebibliography}
\end{document}